\documentclass[runningheads]{llncs}
\usepackage{graphicx}
\usepackage{epstopdf}
\usepackage{multirow}
\usepackage{makecell}
\usepackage{booktabs}
\usepackage{amsmath}
\usepackage{float}
\usepackage{wrapfig}
\usepackage{hyperref}
\hypersetup{colorlinks=true, linkcolor=blue, anchorcolor=black, citecolor=blue}
\usepackage{bm}
\usepackage{amssymb}
\usepackage{bbding}
\usepackage[utf8]{inputenc}
\usepackage[T1]{fontenc}
\usepackage{lmodern} % 提供可缩放的字体
\usepackage{xcolor} % 支持颜色定义
\begin{document}
\title{DyGASR: Dynamic Generalized Exponential Splatting with Surface Alignment for Accelerated 3D Mesh Reconstruction }
\author{Shengchao Zhao\inst{} \and
Yundong Li\inst{(}\Envelope\inst{)} }
\institute{School of Information Science and Technology, North China University of Technology, Beijing, China \\
\email{sczhao@mail.ncut.edu.cn, liyundong@ncut.edu.cn}} 
\maketitle              % typeset the header of the contribution
\begin{abstract}
Recent advancements in 3D Gaussian Splatting (3DGS), which lead to high-quality novel view synthesis and accelerated rendering, have remarkably improved the quality of radiance field reconstruction. However, the extraction of mesh from a massive number of minute 3D Gaussian points remains great challenge due to the large volume of Gaussians and difficulty of representation of sharp signals caused by their inherent low-pass characteristics. To address this issue, we propose DyGASR, which utilizes generalized exponential function instead of traditional 3D Gaussian to decrease the number of particles and dynamically optimize the representation of the captured signal. In addition, it is observed that reconstructing mesh with Generalized Exponential Splatting(GES) without modifications frequently leads to failures since the generalized exponential distribution centroids may not precisely align with the scene surface. To overcome this, we adopt Sugar's approach and introduce Generalized Surface Regularization (GSR), which reduces the smallest scaling vector of each point cloud to zero and ensures normal alignment perpendicular to the surface, facilitating subsequent Poisson surface mesh reconstruction. Additionally, we propose a dynamic resolution adjustment strategy that utilizes a cosine schedule to gradually increase image resolution from low to high during the training stage, thus avoiding constant full resolution, which significantly boosts the reconstruction speed. Our approach surpasses existing 3DGS-based mesh reconstruction methods, as evidenced by extensive evaluations on various scene datasets, demonstrating a 25\% increase in speed, and a 30\% reduction in memory usage.

\end{abstract}
\section{Introduction}
In 3D computer vision, reconstructing surface mesh from multiple calibrated views is a foundational task. Initially, point clouds are typically derived from image collections using traditional Multi-View Stereo (MVS) techniques, and triangular meshes are subsequently constructed from them \cite{kazhdan2013screened}. Recently, Neural Radiance Fields (NeRF) \cite{mildenhall2021nerf} represent neural implicit surface reconstruction techniques and have emerged as formidable competitors. This technique commonly employs Multi-Layer Perceptrons or hash encoding technologies \cite{muller2022instant} to attribute geometric properties like density \cite{ueda2022neural} or the signed distance functions to the nearest surface (SDF) \cite{wang2021neus,yariv2023bakedsdf,li2023neuralangelo} to spatial points. In the domain of neural implicit mesh reconstruction, SDF-based methods are particularly prominent because they facilitate volumetric rendering and enable the learning of implicit surface representations through density functions derived from SDF. Impressive results are attained in small scenes through rendering supervision methods based on neural implicit techniques, however, these methods struggle in complex or large-scale scenes, particularly in those with extensive untextured areas \cite{yu2022monosdf,zhang2022critical}.

To address these challenges, structural priors such as depth \cite{niemeyer2022regnerf}, normal regularization \cite{yu2022monosdf}, point clouds \cite{zhang2022critical}, and semantic information have been incorporated into the optimization process in previous studies, alongside refined sampling strategies such as voxel keypoint guidance \cite{sun2022neural} and hierarchical sampling \cite{zhang2022critical}. While these strategies enhance the accuracy of surface mesh reconstruction, they significantly increase computational demands and extend training duration. Although some methods \cite{zhang2022critical} use MVS-predicted point clouds as priors for mesh reconstruction, these clouds are sparse and noisy, failing to capture the scene's detailed features. Following NeRF, the 3DGS method was recently introduced and has gained popularity \cite{kerbl20233d}. This method excels in generating dense geometric point clouds and explicitly storing the scene's structure in parameter space, enabling direct edits to 3D scenes. Each Gaussian's parameters include position in 3D space, covariance matrix, opacity, and spherical harmonics coefficients. However, Gaussians optimized through 3DGS are not directly usable as priors for mesh reconstruction due to their large numbers, slow training speeds, and predominantly internal scene positioning, which may produce noisy outcomes. 

\begin{wrapfigure}[16]{r}{0.5\textwidth}
    \centering
      \includegraphics[width=.7\linewidth]{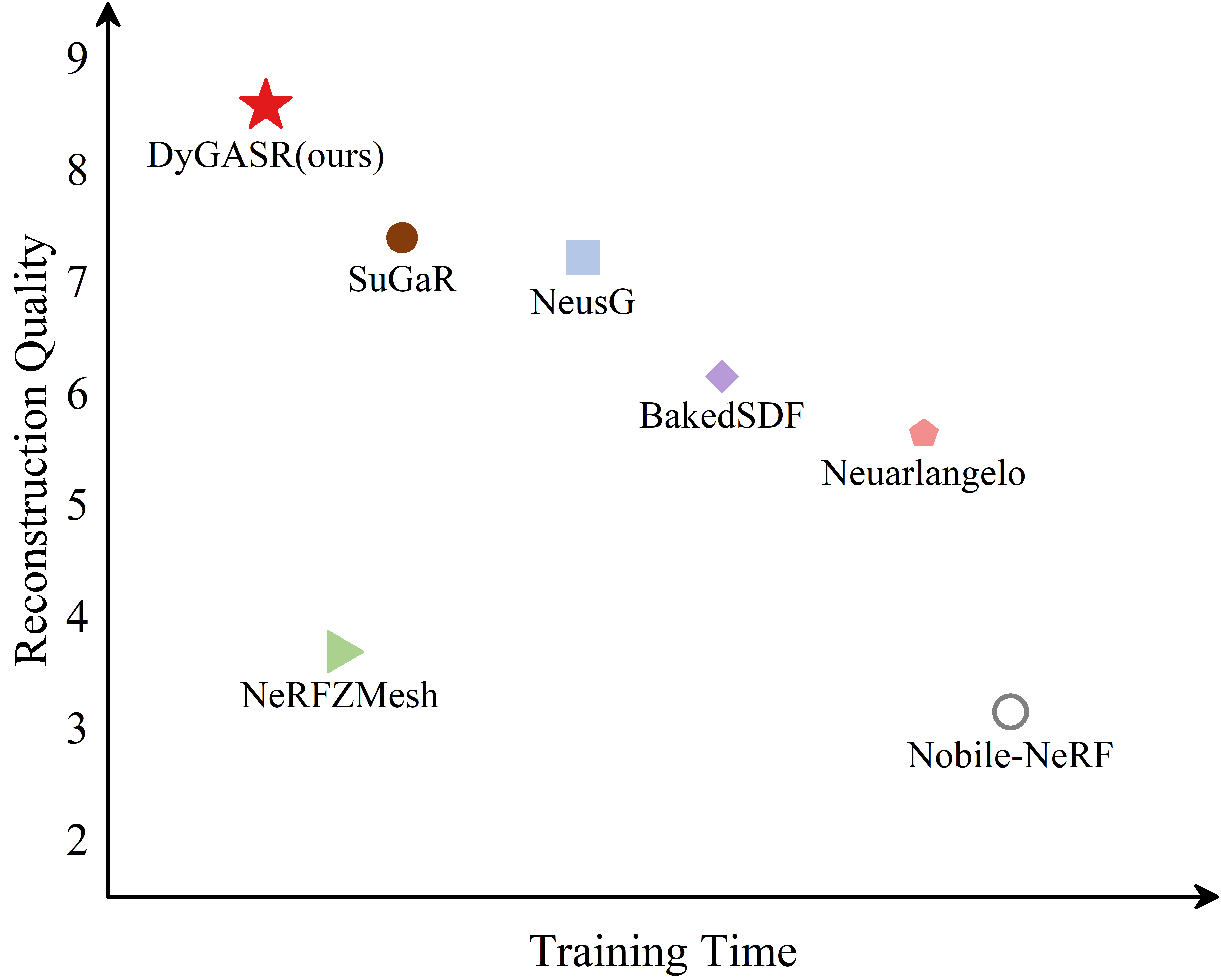}
     \caption{Illustrates that our method, excels in both training time and reconstruction quality, achieving the highest performance.\label{f1}}
\end{wrapfigure}

\begin{figure}[H]
    \centering
    \includegraphics[width=.7\linewidth]{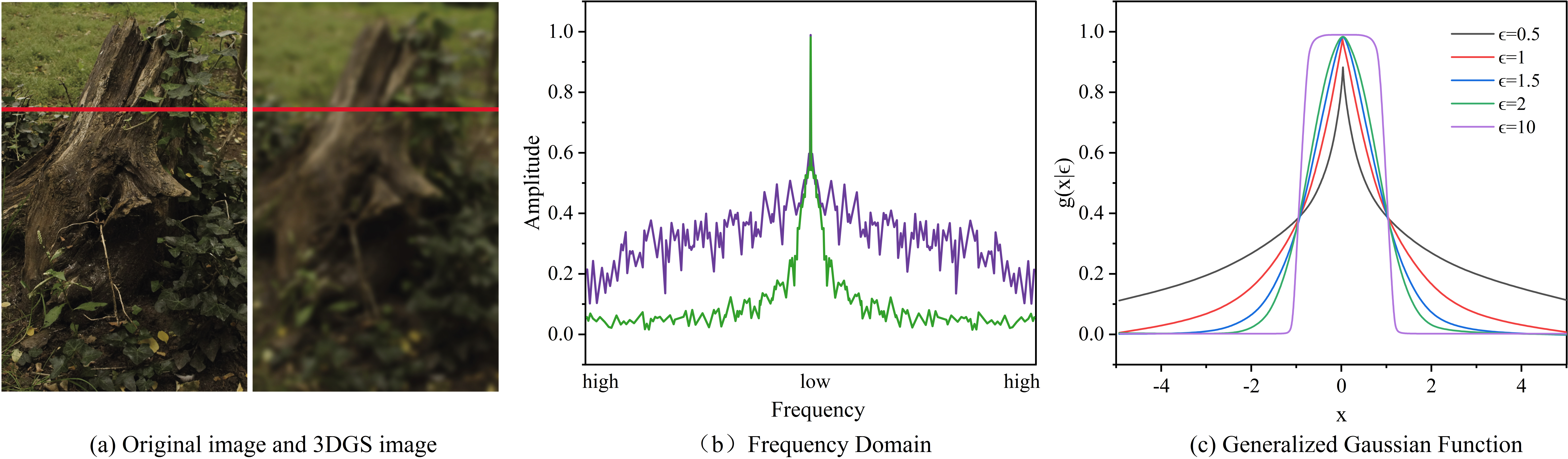}
    \caption{We verify the low-pass characteristics of 3D Gaussians. In (a), the GT image is shown on the left, and a rendering after 500 iterations of 3DGS on the right, both analyzed in the frequency domain via Fourier transform along the same horizontal line. The rendering appears in green and the GT image in blue, highlighting that the low-pass characteristics of 3DGS do not perfectly align with the scene's signal features. In (c), generalized exponential functions are displayed, where $\epsilon=1$ represents a scaled Laplace distribution, $\epsilon=2$, a scaled Gaussian distribution, along with other signal shapes such as triangles and squares. Here, $\epsilon$ serves as a parameter for each component in our method.}
    \label{f2}
\end{figure}

Our method aims to reduce training times and storage costs while surpassing state-of-the-art reconstruction quality. We noted the inherent assumption of low-pass characteristics in 3DGS signal modeling, as illustrated in Figure \ref{f2}(a, b). Given the high-frequency discontinuities in most scenes and the memory burden from numerous tiny Gaussians in 3DGS, we chose a more appropriate basis. Inspired by the work presented in the GES\cite{hamdi2024ges}, we have incorporated the Generalized Exponential Splatting(GES), which represents various signal points with fewer particles and greater precision, over ordinary Gaussians. However, since GES-generated geometric point clouds' centers do not align with actual scene surfaces, we adopt Sugar's approach and introduce Generalized Surface Regularization(GSR)\cite{guedon2023sugar} to align these point clouds with the surface, optimizing parameters to control the shape of generalized exponential splatting. concurrently. Ideally, when generalized exponential splatting distributions are flat and uniformly distributed over the surface, the SDF calculated by the density function minimizes discrepancies with the SDF from actual generalized exponential distribution,  allowing precise representation of surface attributes. A level set, extracted from the density function, undergoes point sampling to facilitate surface mesh creation via Poisson reconstruction \cite{kazhdan2013screened}. Furthermore, we abandoned the original training methods and introduced a strategy that smoothly transitions resolution from small to large, significantly speeding up training convergence and stability while enhancing reconstruction quality. Fig.\ref{f1} displays the high-quality results of our reconstruction and rendering. This work makes the following contributions: (1) We propose DyGASR, employing GES over 3DGS to generate fewer, more effective point cloud priors and introducing GSR to align these priors with scene surfaces, thus accelerating the reconstruction of high-quality surface mesh. (2) Additionally, we apply a dynamic resolution training strategy that smoothly transitions from low to high resolution, effectively shortening training durations and reducing memory consumption. (3) The effectiveness of our method for 3D mesh reconstruction is demonstrated through ablation studies. Evaluations across multiple datasets show a 25\% reduction in training time, a 30\% decrease in memory usage, and a 0.29 dB improvement in PSNR over the SOTA method.

\section{Related work}
\subsection{Multi-view Mesh reconstruction}
\newpage
Multi-view mesh reconstruction is a critical, longstanding area in 3D computer vision reconstruction. Traditional multi-view methods fall into two main categories: point-plane geometric construction and voxel representation. The former method, photometric consistency between images is leveraged to estimate depth maps pixel by pixel, which are then fused to form a dense global point cloud. Techniques like Poisson surface reconstruction delineate the object’s surface structure. However, these methods require high precision in matching and often yield significant artifacts and partial losses in texture-deficient objects. Volumetric reconstruction methods \cite{broadhurst2001probabilistic} adopt an alternative approach that relies on voxel state estimates and avoids direct matching, but they are limited by voxel resolution. Since the incorporation of convolutional neural networks, these methods have been refined for optimization \cite{zhang2020visibility}.

\subsection{Novel View Synthesis}
The new view synthesis method renders an entire 3D scene from a 2D image captured from any viewpoint. NeRF \cite{mildenhall2021nerf} relies on a large multi-layer perceptron to map 3D spatial positions to color and volume density, thereby achieving realistic scene synthesis. However, the use of a large MLP for scene capture leads to extended training times. The introduction of Mip-NeRF \cite{barron2021mip} incorporating improved sampling strategies that enhanced the rendering effects, though the issue of prolonged time consumption persisted. InstantNGP \cite{muller2022instant} combines hash grids and compact MLPs to effectively address the long training time bottleneck. A recent significant advancement in 3DGS \cite{kerbl20233d} utilizes 3D Gaussian point clouds equipped with spatial positions, opacity, covariance matrices, and spherical harmonics, offering a more precise and rapid method of scene acquisition. This method has been rapidly adopted across various domains in computer vision \cite{yan2024street,zielonka2023drivable,yu2023mip}. 3DGS does not heavily depend on neural networks, making it well-suited for scene surface reconstruction. In our work, the original 3DGS is not directly used for reconstructing surface mesh, instead, we streamline it into new Gaussian particles for further operations.

\subsection{Neural Mesh Reconstruction}
A series of precise neural surface reconstruction techniques \cite{yariv2020multiview,wang2022hf,wang2021neus,yariv2023bakedsdf,li2023neuralangelo} aims to deliver high accuracy in surface mesh reconstruction and new viewpoint synthesis. In IDR \cite{yariv2020multiview}, an MLP representing SDF is used to co-learn color and geometric features. Based on this, Neus \cite{wang2021neus} employs weighted volumetric rendering techniques to minimize geometric distortions, while HF-Neus \cite{wang2022hf} decomposes the implicit SDF into base and displacement functions to progressively refine the capture of fine geometric details. Although Neuralangelo \cite{li2023neuralangelo} achieves industry-leading surface reconstruction results with a 3D hash encoding grid, reconstructing a scene can take several days. The recently introduced NeuSG \cite{chen2023neusg}, based on 3DGS, employs a synchronized optimization mechanism combining 3D Gaussians with an implicit SDF neural network, effectively extracting the object's surface mesh. By contrast, our method eschews approaches based on implicit neural radiance fields and avoids joint training with a large number of 3DGS, unlike NeuSG \cite{chen2023neusg}.
\begin{figure}[H]
    \centering
    \includegraphics[width=.76\linewidth]{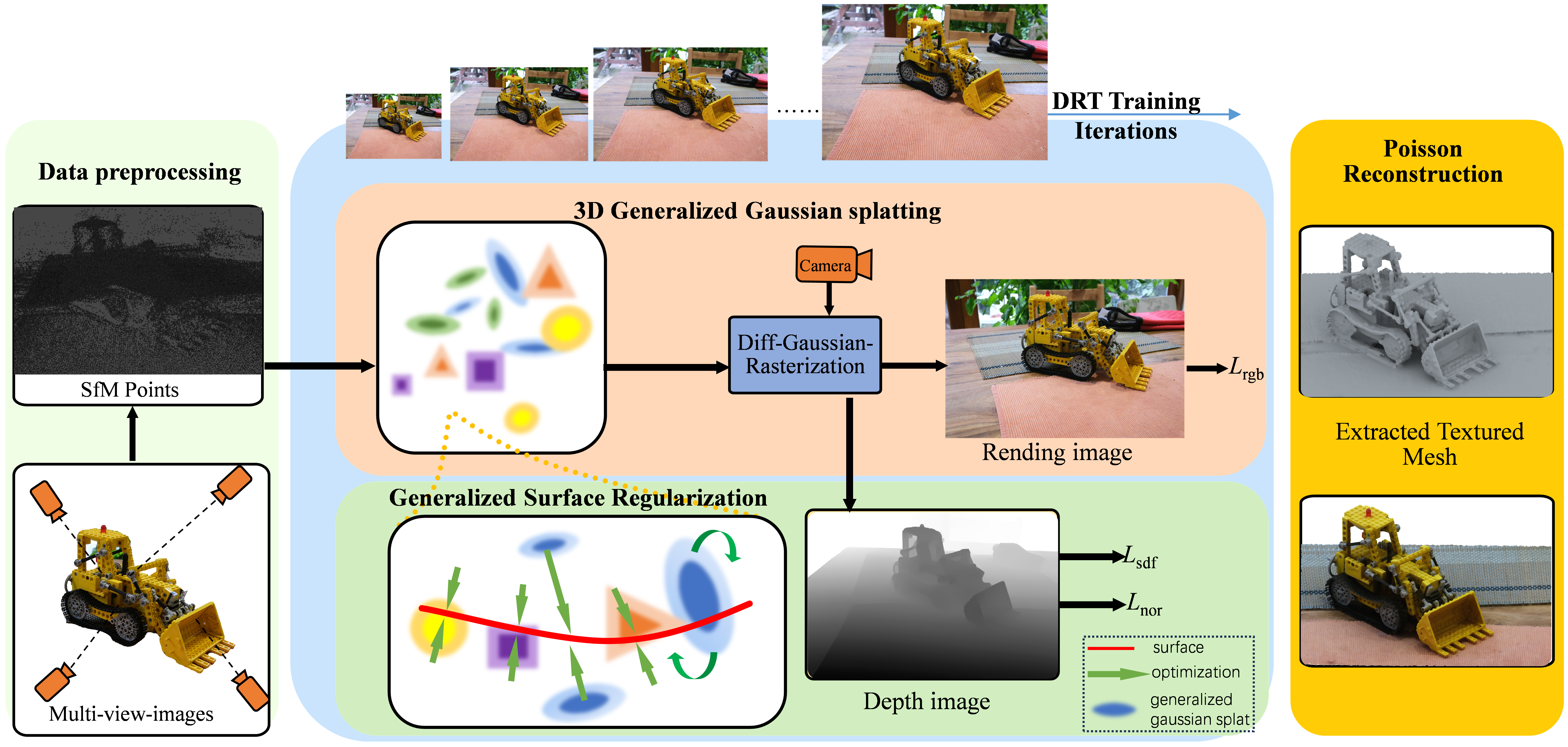}
    \caption{Overview of Our Proposed Framework for Accelerated 3D Mesh Reconstruction.}
    \label{f3}
\end{figure}

\section{Method}
\subsection{Overview\label{s3-1}}

Figure \ref{f3} demonstrates our proposed DyGASR framework. Initially, the Gaussian point  clouds distribution is initialized with sparse point clouds produced via SfM. The method includes three key training components: The GES, described in Section \ref{s3-2}, supervises rendering via projection and rasterization throughout the training process to generate dense generalized exponential point clouds. Notably, many centers of the generated Generalized Exponential Splatting are located within the surface; Thus, the collaborative GSR optimization, as detailed in Section \ref{s3-3}, is integrated to address discrepancies between actual and ideal SDF values, ensuring perpendicularity of actual normals to the surface. This process flattens the 3D generalized exponential splatting and aligns them with the surface; Moreover, the novel DRT training strategy, as introduced in Section \ref{s3-4}, dynamically refines image resolution training from coarse to fine. Following training, the resulting generalized exponentia point cloud is used for Poisson reconstruction to create a textured surface mesh of the scene.

\subsection{Generalized  Exponential Splatting(GES)\label{s3-2}}
To accelerate mesh reconstruction from 3D point clouds using Gaussian Splatting, and inspired by the work of GES\cite{hamdi2024ges}, we adopt GES framework into our 3D surface mesh reconstruction method. This method rasterizes generalized exponential ellipsoids onto images. Utilizing the principles of the Generalized Exponential Function (GEF), GES permits flexible adjustment of generalized exponential primitives by manipulating the shape parameter $\epsilon$. This transformation is shown in Figure \ref{f2}(c). The function is defined as follows:
\begin{equation}
\mathrm{g}(x \mid \delta, \gamma, \epsilon, \mathrm{A})=A \exp \left(-\left(\frac{|x-\delta|}{\gamma}\right)^{\epsilon}\right) 
\end{equation}

Here, $\delta \in \mathbb{R}$ represents the position parameter, $\boldsymbol{\gamma} \in \mathbb{R}$ the scale parameter, $A \in \mathbb{R}^{+}$the amplitude, and $\boldsymbol{\epsilon}>0$ the shape parameter. Specifically, when $\boldsymbol{\epsilon}=2$, the GEF corresponds to a scaled Gaussian distribution:
\begin{equation}
g(x \mid \delta, \gamma, \epsilon=2, A)=\frac{A}{\gamma \sqrt{2 \pi}} \exp \left(-\frac{1}{2}\left(\frac{x-\delta}{\gamma / \sqrt{2}}\right)^{2}\right)    
\end{equation} 

When extended to the GES framework, its core feature is the definition of the position $x$ in 3D space as:
\begin{equation}
  \mathrm{L}\left(x ; x_{\mathrm{i}}, \Sigma, \epsilon\right)=\exp \left\{-\frac{1}{2}\left(\left(x-x_{\mathrm{i}}\right)^{\top} \Sigma^{-1}\left(x-x_{\mathrm{i}}\right)\right)^{\frac{\epsilon}{2}}\right\}  
\end{equation} 

Here, $x_{\mathrm{i}}$ denotes the center of position, and $\Sigma$ corresponds to the covariance matrix in 3DGS. This matrix can be decomposed into the product of a rotation matrix $R$ and a scaling matrix $S$: \begin{equation}
\Sigma=\operatorname{RSS}^{\mathrm{T}} \mathrm{R}^{\mathrm{T}}
\end{equation}

For 2D projections, the covariance matrix $\Sigma^{\prime}$ is computed as $J W \Sigma W^{\mathrm{T}} \mathrm{J}^{\mathrm{T}}$, using the projection matrix $W$ and its Jacobian $J$. However, to maintain $\Sigma$ semi-positive definiteness and adapt $\epsilon$ for the rasterization framework, $S$ is optimized via $\phi(\epsilon)$. As NeRF \cite{mildenhall2021nerf} outlines, volumetric rendering involves calculating expected color through an integration process as light traverses the scene. This is expressed by an integral equation between the light path's near and far boundaries, defining the expected color as:
\begin{equation}
C(r)=\int_{t_{n}}^{t_{f}} T(t) \kappa(r(t)) c(r(t), d) d t, \text {where } T(t)=\exp \left(-\int_{t_{n}}^{t} \kappa(r(s)) d s\right)  
\end{equation}

Here, $T(t)$ denotes transmittance from $t_{n}$ to $t, \kappa(r(t))$ represents volume density, and $c(r(t), d)$ indicates emitted radiance at $r(t)$ in direction $\mathrm{d}$. The cumulative distance $\left[t_{n}, t_{f}\right]$ through nonempty space affects energy attenuation, thereby modulating rendered color intensity. In 3DGS, this distance requires integrating the variance $\gamma$ of projection components along the light ray's direction. In GES, adjustments to the shape parameter influence formula(5) by modulating the effective variance of the projected component $\hat{\gamma}$ of the scaling matrix, controlled by adjustment function $\phi(\boldsymbol{\epsilon})$, defined as follows:
\begin{equation}
    \hat{\gamma}(\boldsymbol{\epsilon})=\phi(\epsilon) \gamma
\end{equation}
\begin{equation}
   \phi(\boldsymbol{\epsilon})=\frac{2}{1+e^{-(\rho \boldsymbol{\epsilon}-2 \boldsymbol{\epsilon})}} 
\end{equation} 

Here, $\rho$ is the shape strength parameter, mitigating potential errors from view-dependent boundary effects and ensuring transformation continuity across different $\epsilon$ values. Finally, the corresponding image projection and rasterization processes are executed to compute the reconstruction loss:
\begin{equation}
  \mathcal{L}_{\text {rgb }}=(1-\lambda) \mathcal{L}_{1}+\lambda \mathcal{L}_{\text {SSIM }} 
\end{equation} 

Here, $\lambda$ is set at 0.2 , allowing GES to continuously optimize $\epsilon$ through rendering loss, utilizing diverse GEF shapes to depict the scene. This approach not only transmits low-frequency signal features but also ensures comprehensive 3D scene coverage while reducing the number of GES. By consistently computing rendering loss globally, the framework infers a more suitable dense point cloud, enhancing mesh reconstruction efficiency.

\subsection{Explicit Mesh Reconstruction\label{s3-3}}
\subsubsection{Generalized Surface Regularization for GES}
Drawing on A.Guédon's SuGaR method\cite{guedon2023sugar}, we integrate it with GES. In the generalized exponential splash scene, considering the corresponding density function $d(x): \mathbb{R}^{3} \rightarrow \mathbb{R}_{+}$at a given position $x$. This function is influenced by the flexible shape parameter $\boldsymbol{\epsilon}$ of the GES model. The function value at $\mathrm{x}$ is computed by summing the values at $x$, each weighted according to the alpha-blending coefficients:
\begin{equation}
 \mathrm{d}(x)=\sum_{\mathrm{i}} \alpha_{\mathrm{i}} \exp \left\{-\frac{1}{2}\left(\left(x-x_{\mathrm{i}}\right)^{\top} \Sigma_{\mathrm{i}}^{-1}\left(x-x_{\mathrm{i}}\right)\right)^{\frac{\epsilon}{2}}\right\}   
\end{equation} 

In an ideal scenario, where GES are perfectly aligned and uniformly distributed, the density at point $x$ is predominantly influenced by the nearest point $g^{*}$, with other points disregarded as appropriate. Additionally, it is essential that the generalized exponential shapes are flat to ensure that they can adhere tightly to the surface even when extremely thin. Consequently, each GES's rotation matrix $\mathbf{s}=\left(s_{1}, s_{2}, s_{3}\right)^{\top}$ must have a minimal $s_{\text {min }}$ factor approaching zero, while the opacity $\alpha_{i}$ is set to 1 . This approach enables a primary focus on the contribution from $g^{*}$ at $\mathrm{x}$, resulting in a simplified expression for the density.
\begin{equation}
    \bar{d}(x) \approx \exp \left(-\frac{1}{2 s_{\text {min }}^{2}}\left\langle x-x_{g^{*}}, n_{g^{*}}\right\rangle^{\epsilon}\right)
\end{equation} 

Here $s$ represents the dimensions of each ellipsoid in every direction, $s_{\min }$ is identified as the minimum scaling factor for the generalized exponential shape, and $n_{g^{*}}$ serves as the principal normal vector of the nearest generalized exponential distribution. Accordingly, densities under both ideal and actual conditions could be integrated into a regularization term. However, empirical evidence indicates that loss calculated using the SDF is more effective than those based on density \cite{guedon2023sugar}, which leding to the incorporation of the SDF expression:
\begin{equation}
    f(x)= \pm s_{\min } \sqrt{-2 \log (d(x))}
\end{equation} 

By substituting the values of $\bar{d}(x)$ for the ideal scenario and $d(x)$ for the actual scenario, the first loss term is derived:
\begin{equation}
    \mathcal{L}_{\text {sdf }}=\frac{1}{|X|} \sum_{x \in X}|\bar{f}(x)-f(x)|
\end{equation} 

$\mathrm{X}$ represents a set of sampled points from key areas within the generalized exponential distribution scene. It has been observed that some sampled points exhibit high gradient values in $\mathcal{L}_{\text {sdf }}$, and the normals at $g^{*}$ points are perpendicular to the surface. Consequently, a second regularization loss is introduced, aligning the normals in the actual state more closely with $n_{g^{*}}$ :
\begin{equation}
    \mathcal{L}_{\text {nor }}=\frac{1}{|X|} \sum_{x \in X}\left\|\frac{\nabla f(x)}{\|\nabla f(x)\|_{2}}-n_{g^{*}}\right\|_{2}^{2}
\end{equation} 

Here $\nabla f(x)$ is the gradient of the SDF computed at point $x$. Consequently, the total loss function of our model is defined as follows:
\begin{equation}
    \mathcal{L}_{\text {total }}=\mathcal{L}_{\text {rgb }}+\lambda_{1} \mathcal{L}_{\text {sdf }}+\lambda_{2} \mathcal{L}_{\text {nor }}
\end{equation} 

In this formulation, $\lambda_{1}$ and $\lambda_{2}$ are used to weight the two regularization terms. Through concerted optimization, the dense point clouds produced by the GES are precisely flattened and aligned.

\subsubsection{Mesh Extraction}
To rapidly generate mesh from regularized GES, the Poisson reconstruction algorithm is employed \cite{kazhdan2013screened}. An isosurface, depending on parameter $\alpha$, is defined by sampling 3D point sets on a density function determined by Gaussians. This process involved randomly selecting pixel points from the generalized exponential distribution depth map as origins for line-of-sight directions, where the depth map is obtained using an extended splatting rasterizer. Sampling along each selected pixel's line of sight direction $v$, points $x+t_{k} v$ are generated. The range for $t_{k}$, set to $\left[-3 \sigma_{g}(v), 3 \sigma_{g}(v)\right]$ is based on the standard deviation $\sigma_{g}(v)$ of the generalized exponential distribution $g$ in direction $v$, covering the $99.7 \%$ confidence interval. Density values $d_{k}=d\left(x+t_{k} v\right)$ at each sample are calculated, and intervals satisfying $d_{k}<\lambda<d_{l}$ are identified to locate isosurface points. Using linear interpolation, the isosurface point closest to the camera $x+t^{*} v$ is determined to satisfy $d(x+$ $\left.t^{*} v\right)=\alpha$. At each isosurface point $\hat{x}$, the surface normal is calculated, defined by the normalized gradient of the density function $\frac{\nabla f(x)}{\|\nabla f(x)\|_{2}}$. Subsequently, the mesh is reconstructed using the Poisson algorithm, utilizing these isosurface points and their normal information. After the initial mesh extraction, the mesh is further refined by binding new 3D generalized exponential distribution to the mesh triangles and co-optimizing. This process employs a Gaussian rasterizer, allowing for the editing of Gaussian-flattened scenes using common mesh editing tools while maintaining high-quality rendering effects.

\subsection{Dynamic Resolution Training\label{s3-4}}
Traditional Gaussian training is uniformly conducted at a single resolution throughout the entire image, resulting in a suboptimal loss landscape. Consequently, we introduce a Dynamic Resolution Training (DRT) strategy, employing a coarse-to-fine approach to transforming the conventional training paradigm. Training initially begin at a low resolution and progressively increased to full resolution according to a cosine schedule, the downsampling scaling factor is defined as:
\begin{equation}
\text{value} = \text{end} +0.5 \cdot(\text{ start - end }) \cdot\left(1+\cos \left(\frac{\text { current\_iteration }}{\text { total\_iteration }} \cdot \pi\right)\right)    
\end{equation}

Here, start and end denote the strategy's initial and final positions, respectively, with current\_iteration and total\_iteration indicating the counts of current and total iterations. Initially, utilizing sparse point clouds and approximate attributes, early optimization of details hindered convergence and risked Gaussian blur artifacts. As the resolution increased, it allowed for better accommodation of generalized exponential distribution, thereby enhancing the reconstruction of fine features. In summary, the DRT strategy significantly diminished training durations while beneficially influencing the reconstruction quality of scenes.
\begin{figure}[H]
    \centering
    \includegraphics[width=\linewidth]{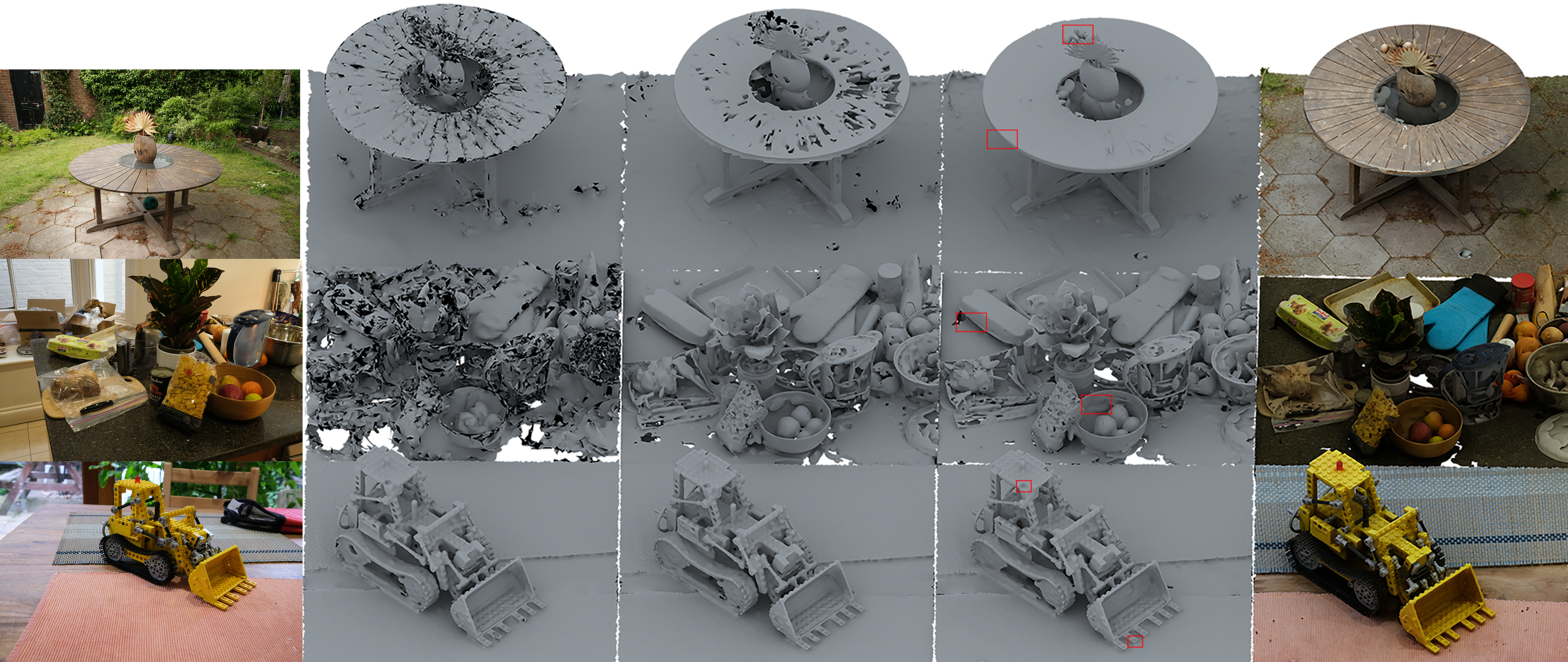}
    \begin{minipage}{0.19\linewidth}
    \centering
  \footnotesize  Ground Truth 
     \end{minipage}
         \begin{minipage}{0.19\linewidth}
         \centering
      \footnotesize 	 NeRF2Mesh	
      \end{minipage}
         \begin{minipage}{0.19\linewidth}
         \centering
      \footnotesize     SuGaR	
          \end{minipage}
             \begin{minipage}{0.19\linewidth}
             \centering
      \footnotesize         Ours
              \end{minipage}
                 \begin{minipage}{0.19\linewidth}
                 \centering
             \footnotesize      Ours(color)
    \end{minipage}
    \caption{Qualitative Comparison of Three Scenes from the Mip-NeRF360 Dataset \cite{barron2021mip}. The Last Column Displays Results from Our Rendering Method.}
    \label{f4}
\end{figure}

\section{Experiments}
\subsection{Datasets and Metrics}
To assess the performance of our method, experiments are carried out on seven scenes from the Mip-NeRF360 dataset \cite{barron2021mip} and two scenes from the DeepBlending dataset \cite{hedman2018deep}, encompassing 200 to 630 RGB images. Before analysis, images are preprocessed using the Colmap algorithm to initialize sparse scene point clouds. In this research, PSNR, SSIM, and LPIPS \cite{zhang2018unreasonable} are employed as evaluation metrics to gauge the rendering quality obtained with the reconstructed mesh and its associated surface generalized exponential distribution.

\subsection{Implementation}
Our method employs the PyTorch framework and CUDA backend for rasterization operations, with training performed on a single Nvidia RTX 3090.

\textbf{Generalized Exponential Splatting:} To accurately locate the 3D generalized exponential shape, iterative optimization is conducted continuously, with the state parameter's learning rate set at $1.5 \mathrm{e}-3$, the shape reset interval set at 1000 iterations, the shape pruning threshold set at 0.5, and the intensity parameter $\rho$ set at 0.1. Other hyperparameters remain consistent with 3DGS standards.

\textbf{Regularization:} After 7000 iterations, the opacity $\alpha_{\mathrm{j}}$ is subjected to 1800 entropy regularizations to enforce a binary state. Across 6200 iterations, the GSR technique outlined in Section 3.3 is applied, and generalized exponential distribution with opacity below 0.5 are eliminated. For the density values of generalized exponential points $g^{*}$, sums are calculated from $x_{i}^{\prime}$ 's 16 closest points using KNN algorithm, with the nearest neighbor list updated every 500 iterations.

\textbf{Mesh extraction:} An isosurface with $\alpha=0.3$ is selected, and reconstruction is performed using a depth-10 Poisson method, followed by 15,000 iterations to co-optimize the mesh and the associated generalized exponential distribution.

\textbf{Dynamic Resolution Training:} Throughout $75 \%$ of the process, scaling operations are executed, beginning at a scaling factor of 0.26 and incrementally increasing to 1 via a cosine schedule.

\subsection{Main Results}

\begin{table}[H]
    \centering 
    \caption{ Comparison of our method with other leading approaches across two datasets and nine scenes, demonstrating the optimal performance of our method.}
\resizebox{\linewidth}{!}{
\begin{tabular}{cccccc|ccccc}
\hline
\multirow{2}{*}{}\begin{tabular}{c}
Datasets 
\end{tabular} & \multicolumn{5}{c|}{Mip-NeRF360\cite{barron2021mip}} & \multicolumn{5}{c}{Deep Blending\cite{hedman2018deep}} \\
\cline{2-11}
 Method  & PSNR$\uparrow$ & SSIM$\uparrow$ & LPIPS$\downarrow$ & \begin{tabular}{c}
TrainingTime \\
(hour)$\downarrow$ \\
\end{tabular} & \begin{tabular}{c}
VRAM \\
$(\mathrm{GB})$$\downarrow$ \\
\end{tabular} & PSNR$\uparrow$ & SSIM$\uparrow$ & LPIPS$\downarrow$ & \begin{tabular}{c}
TrainingTime \\
(hour)$\downarrow$ \\
\end{tabular} & \begin{tabular}{l}
VRAM \\
(GB)$\downarrow$ \\
\end{tabular} \\ \hline
Mobile-NeRF\cite{chen2023mobilenerf} & 22.95 & 0.582 & 0.393 & $>180$ & 24 & 23.52 & 0.573 & 0.385 & $>120$ & 24 \\
BakedSDF\cite{yariv2023bakedsdf} & 26.77 & 0.711 & 0.304 & $>90$ & 24 & 25.79 & 0.786 & 0.314 & $>60$ & 24 \\
NeRF2Mesh\cite{tang2023delicate} & 23.04 & 0.523 & 0.457 & 1.42 & 15 & 23.96 & 0.594 & 0.367 & 0.83 & 9 \\
neuarlangelo\cite{li2023neuralangelo} & 25.68 & 0.681 & 0.354 & $>124$ & 24 & 26.56 & 0.805 & 0.326 & $>90$ & 24 \\
NeuSG\cite{chen2023neusg} & 26.98 & 0.756 & 0.289 & 16 & 24 & 27.34 & 0.811 & 0.332 & 10 & 24 \\
sugar\cite{guedon2023sugar}] & 27.28 & 0.813 & 0.262 & 1.61 & 20.4 & 27.88 & 0.836 & 0.34 & 0.98 & 10.5 \\
DyGASR (ours) &\bf 27.57 &\bf 0.831 &\bf 0.248 &\bf 1.25 &\bf 14 &\bf 29.05 &\bf 0.881 &\bf 0.269 &\bf 0.76 &\bf 8.5 \\
\hline
\end{tabular}}
    \label{t1}
\end{table}

Table \ref{t1} showcases the qualitative analysis of our method compared with several leading-edge approaches. When contrasted with neural implicit reconstruction techniques \cite{yariv2023bakedsdf,li2023neuralangelo,chen2023mobilenerf,tang2023delicate}, our method achieved a speed increase of approximately $199 \%$ and a $42 \%$ reduction in VRAM usage. Compared to the prevalent and efficient 3DGS-based methods \cite{chen2023neusg,guedon2023sugar}, our approach realized an $85 \%$ improvement in speed and a $37 \%$ decrease in VRAM usage, while enhancing reconstruction quality. Particularly with the state-of-the-art SuGaR \cite{guedon2023sugar} method, our method not only enhanced speed by $25 \%$ and reduced GPU RAM consumption by $6.3 \mathrm{~GB}$, but also excelled in detailed reconstruction. For instance, as depicted in Figure \ref{f4}, our approach yielded a more comprehensive and detailed surface mesh. In the 'garden' scene, the desktop displayed virtually no gaps, with the table edges and vase maintaining an orderly structure. In the 'counter' scene, our method exhibited significantly less noise than previous approaches. Although the reconstruction in the 'kitchen' is commendable across methods, our method retained more precise details, such as the finer features of the tongs.

\subsection{Ablations Study}
\subsubsection{Quantity and Quality of generalized exponential point clouds} 
\begin{table}[H]
    \centering
   \caption{Qualitative and quantitative ablation study on the number of point clouds and their effects on quality, utilizing the bicycle scene data. Various components are used to reduce the number, illustrating their effects on the reconstruction quality.}
\footnotesize
\begin{tabular}{ccc|cc}
\hline
\begin{tabular}{c}
Normal \\
3DGS\\
\end{tabular} & \begin{tabular}{c}
GES\\
\end{tabular} & DRT & Number $\downarrow$ & PSNR $\uparrow$ \\
\hline
$\checkmark$ & $\times$ & $\times$ & $2.78 \mathrm{M}$ & $23.04 \mathrm{~dB}$ \\
$\times$ & $\checkmark$ & $\times$ & $1.76 \mathrm{M}$ & $23.01 \mathrm{~dB}$ \\
$\checkmark$ & $\times$ & $\checkmark$ & $2.63 \mathrm{M}$ & $23.15 \mathrm{~dB}$ \\
$\times$ & $\checkmark$ & $\checkmark$ & $1.72 \mathrm{M}$ & $23.18 \mathrm{~dB}$ \\
\hline
\end{tabular}
    \label{t2}
\end{table}

As Table \ref{t2} shows, substituting ordinary 3DGS with GES resulted in a slight decrease in quality, but reduced the number of point clouds by 1.02 million significantly. Furthermore, integrating the DRT strategy achieved an optimal balance between the number of point clouds and their quality. These results confirm that our method not only eliminates redundant particles and optimizes particle distribution but also increases processing speed and reduces memory consumption.

\subsubsection{Loss using GSR and DRT} 

\begin{figure}[H]
    \centering
    \includegraphics[width=.62\linewidth]{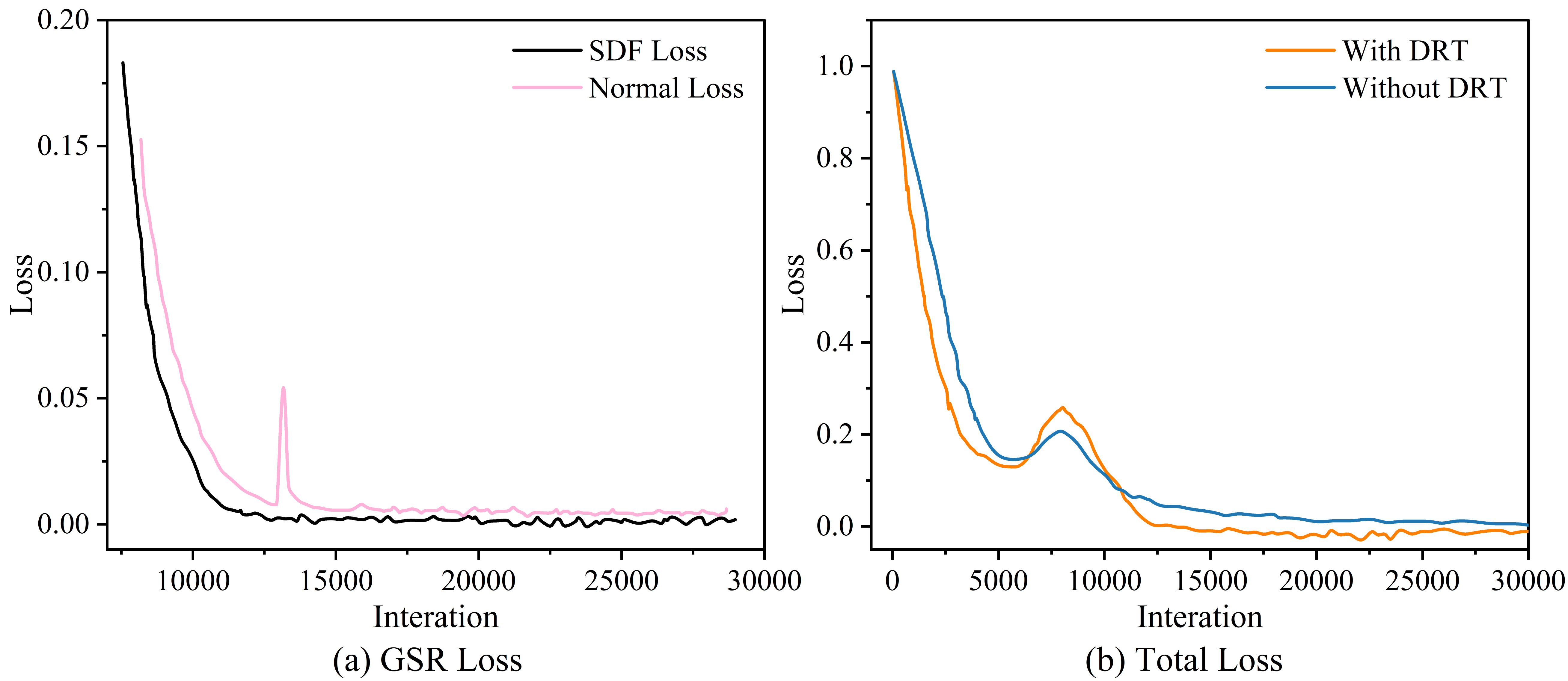}
    \caption{Loss curves averaged from three random seeds in the bonsai scene. On the left are two types of losses in GSR, and on the right are training losses before and after the application of DRT.}
    \label{f5} 
\end{figure}
 
To demonstrate the transformation of 3D generalized exponential distribution into a flat plane, Figure \ref{f5}(a) displays the GSR, incorporating both SDF and normal losses. The SDF loss approaches zero (approximately $1 \times 10^{-7}$ ), confirming an effective transformation of the 3D generalized exponential distribution into an extremely thin structure. The normal loss approaches zero, indicating that the actual normal directions become effectively perpendicular to the surface. As illustrated in Figure \ref{f5}(b), integrating DRT stabilized the loss landscape and accelerated the model's convergence rate.

\subsubsection{Impacts of different components}
To visually illustrate the effectiveness of DyGASR's components, ablation studies were conducted across Mip-NeRF360 dataset \cite{barron2021mip}, utilizing extracted mesh and their corresponding generalized exponential distribution surfaces to assess DyGASR's rendering quality. Results are shown in Table \ref{t3}, with SuGaR \cite{guedon2023sugar} serving as the baseline. The results indicate that the sole use of GSR is not significantly effective. Changes in Gaussian types and modifications in training methods cut processing times by 11 and 18 minutes, and decreased VRAM usage by $5 \mathrm{~GB}$ and 2GB, respectively. Consequently, the PSNR values increased by $0.19 \mathrm{~dB}$ and $0.15 \mathrm{~dB}$.
\begin{table}[H]
\tabcolsep  0.5cm
    \centering
    \caption{Ablation study of different components within our method, averaged across seven experimental scenes in the Mip-NeRF360 dataset \cite{barron2021mip}.}
    \footnotesize
\begin{tabular}{ccccc}
\hline
Ablation Setup & PSNR $\uparrow$ & TrainingTime $\downarrow$ & VRAM $\downarrow$ \\
\hline
Baseline\cite{guedon2023sugar} & $27.28 \mathrm{~dB}$ & $1.61 \mathrm{~h}$ & $21 \mathrm{G}$ \\
DyGASR w/ GSR & $27.32 \mathrm{~dB}$ & $1.65 \mathrm{~h}$ & $21 \mathrm{G}$ \\
DyGASR w/o GES & $27.38 \mathrm{~dB}$ & $1.43 \mathrm{~h}$ & $19 \mathrm{G}$ \\
DyGASR w/o DRT & $27.42 \mathrm{~dB}$ & $1.55 \mathrm{~h}$ & $16 \mathrm{G}$ \\
Full DyGASR & $27.57 \mathrm{~dB}$ & $1.25 \mathrm{~h}$ & $14 \mathrm{G}$ \\ \hline
\end{tabular}
    \label{t3}
\end{table}

\section{conclusion}
We introduce DyGASR, an innovative approach for accelerating 3D mesh reconstruction using dynamic generalized exponential  splats aligned with surfaces. This method employs Generalized Exponential Splatting model over traditional ones, reducing the number of required particles and enhancing signal feature precision. The integration of a generalized surface regularization module guarantees a more accurate alignment of generalized exponential distribution centroids with actual scene surfaces, thereby enhancing mesh precision. Additionally, our dynamic resolution adjustment strategy markedly accelerates training and cuts memory consumption. Comparative evaluation with advanced 3DGS-based methods demonstrates our method's $25 \%$ speed increase, $30 \%$ memory reduction, and superior quality, establishing a new benchmark in $3 \mathrm{D}$ mesh reconstruction.

% \vspace{\baselineskip} % 添加一行间距
% \noindent{\bfseries Acknowledgements. }This work was supported by the National Natural Science Foundation of China under Grant 62071006. 

\nocite{*}
\bibliographystyle{splncs04}
\bibliography{ref}

\end{document}